\documentclass[10pt,twocolumn,letterpaper]{article}

\usepackage{graphicx}
\usepackage{amsmath}
\usepackage{amssymb}
\usepackage{booktabs}
\usepackage{tikz}
\usepackage{svg}
\usepackage{bm}
\usepackage{array}
\usepackage{makecell}

\usepackage[pagebackref,breaklinks,colorlinks]{hyperref}

\usepackage[capitalize]{cleveref}
\crefname{section}{Sec.}{Secs.}
\Crefname{section}{Section}{Sections}
\Crefname{table}{Table}{Tables}
\crefname{table}{Tab.}{Tabs.}

\begin{document}

\title{Extreme compression of sentence-transformer ranker models:\\ faster inference, longer battery life, and less storage on edge devices\thanks{This work is conducted at Xayn AG, Unter den Linden 42, 10117 Berlin.}}


\author{Amit Chaulwar\\
{\tt\small amit.chaulwar@xayn.com}
\and
Lukas Malik\\
{\tt\small lukas.malik@xayn.com}
\and
Maciej Krajewski \\
{\tt\small Maciej.Krajewski@xayn.com}
\and 
Felix Reichel \\
{\tt\small felix.reichel@xayn.com}
\and
Leif-Nissen Lundbæk \\
{\tt\small leif@xayn.com}
\and
Michael Huth \\
{\tt\small michael@xayn.com}
\and 
Bartlomiej Matejczyk\thanks{This is the corresponding author.}\\
{\tt\small bartlomiej.matejczyk@xayn.com}
}
\maketitle

\begin{abstract}
   Modern search systems use several large ranker models with transformer architectures. These models require large computational resources and are not suitable for usage on devices with limited computational resources. Knowledge distillation is a popular compression technique that can reduce resource needs of such models, where a large teacher model transfers knowledge to a small student model. To drastically reduce memory requirements and energy consumption, we propose two extensions for a popular sentence-transformer distillation procedure: generation of an optimal size vocabulary and dimensionality reduction of the embedding dimension of teachers prior to distillation. We evaluate these extensions on two different types of ranker models. This results in extremely compressed student models whose analysis on a test dataset shows the significance and utility of our proposed extensions.

\end{abstract}

\section{Introduction} \label{sec:introduction}
\label{sec:intro}

Usage of search engines has increased dramatically in  last decades, reflecting the digitization of a vast amount of knowledge, the ubiquity of handheld and connected devices, and the need to access knowledge within seconds.  

Criteria that measure the quality of search experience include information relevance, personalisation, latency, privacy, and energy efficiency.  Many search engines are competing to provide better search experience by focusing only on some of these criteria. For example, some search engines focus on providing personalised and relevant information using large machine learning models with massive computing infrastructure; but this comes at the expense of having to track sensitive and personal data of their users.  

Some search engines focus on privacy, instead, by removing personally identifiable information from search logs; this comes at the expense of poorer personalization of search results. The saved search logs, even when anonymised, pose a risk of re-identification attacks~--~especially when data is leaked or systems are hacked. So far, there seems to be no search engine that provides highly personalised, relevant results with strong user privacy guarantees.

\tikzstyle{block} = [rectangle, draw, fill=blue!20, 
    text width=5em, text centered, rounded corners, minimum height=5em]
\tikzstyle{block1} = [rectangle, draw,text width=5em, text centered, rounded corners, minimum height=3em]

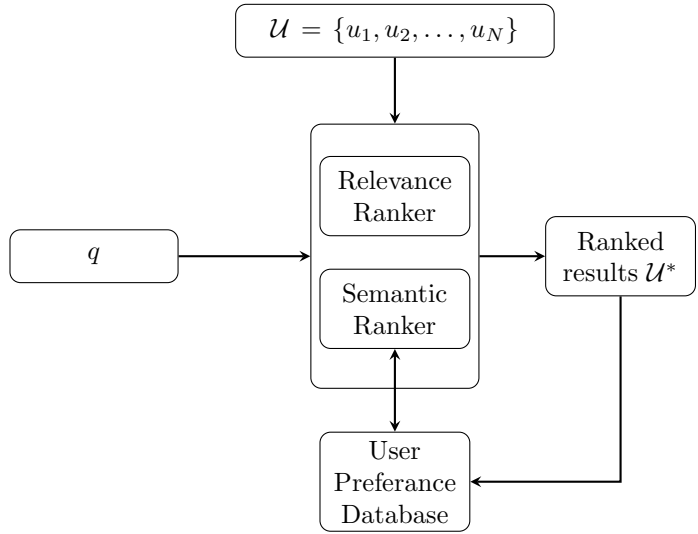
\begin{figure*}[t]
    \centering
    {
    \begin{tikzpicture}
    
     \node (W10) [block1,text width=2 cm,align=center, minimum height=2em] {$q$};
    \node (W11) [block1, right of = W10, text width=2 cm,align=center, minimum height=10em, xshift = 3 cm] {};
    \node (W00) [block1, above of= W11,text width=4 cm,align=center, minimum height=2em, yshift=2 cm] {$\mathcal{U} = \{u_1,u_2,\ldots, u_N\}$};
    \node (W12) [block1, below of=W11, yshift = 0.3 cm]{Semantic Ranker};
    \node (W13) [block1, below of=W11, yshift = 1.8 cm]{Relevance Ranker};
    \node (W14) [block1, right of = W11, xshift= 2 cm] {Ranked results $\mathcal{U}^*$};
    \node(W15) [block1, below of = W11, yshift=-2 cm] {User Preferance Database};
    
    \draw [thick,->,>=stealth] (W10) -- (W11);
    \draw [thick,->,>=stealth] (W00) -- (W11);
    \draw [thick,->,>=stealth] (W11) -- (W14);
    \draw [thick,->,>=stealth] (W14.270) |- (W15);
     
    \draw[thick,<->,>=stealth] (W15) -- (W12);
    \end{tikzpicture}
    }
    \caption{A general design overview for a personalised search system.}
    \label{fig:searchsystem}
\end{figure*}

We stipulate that a good search experience should provide search results that are pertinent to the user query, energy efficient in their generation, rendered with low latency, and equipped with strong privacy guarantees. We resolve the tension between personalisation and privacy of search results through the use of on-device machine learning models. These models providing users with personalised results that are relevant to the user query. This privacy by design ensures that sensitive information can remain on the device controlled by the user and never leave that device.

Natural language processing (NLP) is a central tool used to build modern search engines that offer relevance and personalization of search experiences. The NLP models that render such tools are huge and need to support a range of languages in which a user engages during searches. 

The biggest challenge we face in realizing our vision of a user-centric and privacy-preserving search technology is how to modify huge NLP models so that they can be deployed on consumer devices such as smartphones, without degrading the performance of these devices and the search experience.

The size of the state-of-the-art NLP models is increasing exponentially \cite{simon_2021}. Using such large models in edge devices is impractical as they require too much memory, have unsatisfactory inference time, and consume too much energy. Model compression methods such as distillation \cite{Hinton2015DistillingTK}, quantization \cite{Gholami2022ASO},  and prunning \cite{248452} are therefore widely used.

We increase the utility of such compression methods for better user experience by developing and evaluating an extension for the distillation method proposed in \cite{Reimers2020MakingMS};  this method reduces the size of multilingual NLP models that are used in search engines and other NLP applications beyond search.
The contributions we report in this paper are:
\begin{enumerate}
    \item We prove that generating multilingual vocabulary for a set of target languages instead of using a multilingual vocabulary of state-of-the-art multilingual models trained on hundreds of languages improves the performance without a change of model architecture. 
    \item We show how to reduce the embedding dimension of the teacher model with dimensionality reduction techniques such as PCA to produce a student model with smaller embedding dimensions and significantly reduced size.
    \item We evaluate the utility of our extension on state-of-the-art models to assess its  suitability to edge devices~--~based on criterion such as latency, memory, inference time, and energy efficiency.
    
\end{enumerate}

\textbf{Outline}: Section \ref{sec:search_system_design} describes the type of ranker models used in a general search engine for which model compression techniques are proposed.  Section \ref{sec:background} provides a survey on knowledge distillation, an important model compression technique. This section also describes the training procedure for ranker models which are used for knowledge distillation. The intuition behind our proposed solutions for model compression is presented in Section \ref{sec:proposed_solution}, where a detailed description of proposed methods is provided in Section \ref{sec:distillation_extensions}. The findings of our comparative analysis for different distilled model with the original ranker models
is summarised in Section \ref{sec:comparative_analysis}. The paper concludes and discusses scope for future work in Section \ref{sec:conclusion}.

\textbf{Notation}: Throughout this paper, vectors are denoted by lower case boldface letters.

\section{Search System Design} \label{sec:search_system_design}

A general design for a personalised search system is shown in Fig.\,\ref{fig:searchsystem}. In this section, we illustrate how ranker models are needed and used to support a personalised search system whose models are obtained from the compression techniques proposed in this paper. We omit details of the personalized search system as these are not relevant for the focus of this paper.

  The input to the search  system is a set of $N$ documents $\mathcal{U} = \{u_1,u_2,\ldots,u_N\}$ and a user query $q$. These documents could stem from a sole, large database or they could be fetched from multiple sources or APIs. No matter how these documents are sourced, we want to present these documents to the user in the order $\mathcal{U^*}$ such that top documents are not only more relevant to the user query $q$ but also relevant to the user preferences. These user preferences are captured from user interactions with documents returned for past queries. 
  
  To support such search systems, we need two ranker models:
\begin{enumerate}
      \item Semantic ranker (S-mBERT) for ranking documents according to the user interests and disinterests, 
      \item Relevance ranker (Q-mBERT) for ranking documents according to the relevance to the query.
\end{enumerate}

The semantic ranker should be able to convert user interests into vector representations that can be used for personalised ranking. To guarantee strong user privacy, both ranker models must be   small enough and computationally efficient so that they can run directly on the edge device of the user. This ensures that sensitive, personal data can stay on and never leave that device.

\section{Background} \label{sec:background}
The introduction of Bidirectional Encoder Representations from Transformers (BERT) \cite{devlin2018bert} quickly led to their huge popularity as models of choice for NLP research, products, and services. Following approaches such as GPT \cite{Brown2020LanguageMA} and RoBERTa \cite{Liu2019RoBERTaAR} complement this innovative tool box in NLP that opens up variety of exciting applications. 

These approaches see a dramatic increase in the number of model parameters; the size of these models appears to increase exponentially \cite{simon_2021} with no limit in sight, as the performance of these models is correspondingly improving. 

Most of these models are open-sourced, but their usage is impractical on edge devices where the computational and energy resources are limited. Therefore, the NLP community has investigated several model compression techniques. In this paper, we use the knowledge distillation method in \cite{Hinton2015DistillingTK} for model compression of large sentence-transformer models \cite{Reimers2019SentenceBERTSE}. We refer to this method as  \emph{distillation} subsequently.

Distillation is the process of transferring knowledge from a large machine learning model (the teacher model) to a small machine learning model (the student model). Put simply, the student network is trained to mimic the behaviours of the teacher network. Several approaches have been proposed for the distillation of BERT models \cite{Sun2020MobileBERTAC, Jiao2020TinyBERTDB, Sanh2019DistilBERTAD, Zhao2019ExtremeLM}; but their inference time is still very high and they only distil a pre-trained BERT model which requires fine-tuning for specific tasks. 

The sentence-transformers (SBERT) \cite{Reimers2019SentenceBERTSE} are a modification of the pre-trained BERT network that use siamese network structures to derive semantically meaningful sentence embeddings that can be compared using cosine-similarity. Compared to the pre-trained BERT models, this is more efficient: sentence-embeddings can be stored and test sentence embeddings can be much more efficiently computed and compared with stored sentence embeddings through cosine similarity. 

A distillation procedure for sentence-transformers models has been proposed in \cite{Reimers2020MakingMS};  it not only compresses the models but also converts a monolingual teacher model into a multilingual student model. However, the size of the student model is still too large for use on edge devices. 

\tikzstyle{block} = [rectangle,  
    text width=5em, text centered, rounded corners, minimum height=3em]
\tikzstyle{block1} = [rectangle, draw,text width=5em, text centered, rounded corners, minimum height=3em]

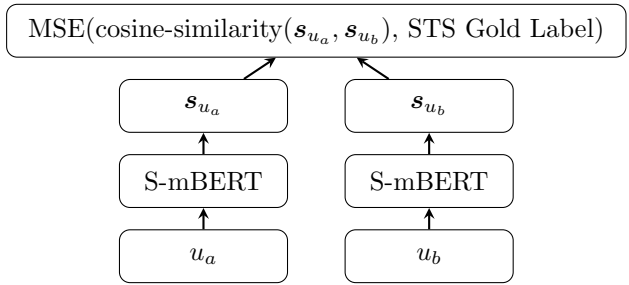
\begin{figure}[t]
    \centering
    
    {
    \begin{tikzpicture}
     \node (W10) [block1, text width=2 cm,align=center, minimum height=2em] {$u_a$};
     \node (W20) [block1, right of = W10,text width=2 cm,align=center, minimum height=2em, xshift = 2 cm] {$u_b$};
     \node (W11) [block1, above of=W10, text width=2 cm,align=center, minimum height=2em] {S-mBERT};
     \node (W21) [block1, above of= W20, text width=2 cm,align=center, minimum height=2em] {S-mBERT};
     \node (W12) [block1, above of = W11, text width=2 cm,align=center, minimum height=2em] {$\bm s_{u_a}$};
     \node (W22) [block1, above of = W21, text width=2 cm,align=center, minimum height=2em] {$\bm s_{u_b}$};
     \node (W30) [block1, above of = W22, text width=8 cm,align=center, minimum height=2em, xshift= -1.5 cm, yshift=0. cm] {MSE(cosine-similarity$(\bm s_{u_a},\bm s_{u_b}),$ STS Gold Label)};

    \draw [thick,->,>=stealth] (W10) -- (W11);
    \draw [thick,->,>=stealth] (W11) -- (W12);
    \draw [thick,->,>=stealth] (W12) -- (W30);
    \draw [thick,->,>=stealth] (W20) -- (W21);
    \draw [thick,->,>=stealth] (W21) -- (W22);
    \draw [thick,->,>=stealth] (W22) -- (W30);

    \end{tikzpicture}
    }
    \caption{Training procedure for semantic ranker teacher model.}
    \label{fig:semantic_ranker_training}
\end{figure}
\tikzstyle{block} = [rectangle,  
    text width=5em, text centered, rounded corners, minimum height=3em]
\tikzstyle{block1} = [rectangle, draw,text width=5em, text centered, rounded corners, minimum height=3em]

\begin{figure}[t]
    \centering
    {
    \begin{tikzpicture}
     \node (W10) [block1, text width=2 cm,align=center, minimum height=2em] {$q$};
     \node (W20) [block1, right of = W10,text width=2 cm,align=center, minimum height=2em, xshift = 2 cm] {$u_p$};
     \node (W30) [block1, right of = W20,text width=2 cm,align=center, minimum height=2em, xshift = 2 cm] {$u_n$};
     \node (W11) [block1, above of=W10, text width=2 cm,align=center, minimum height=2em] {Q-mBERT};
     \node (W21) [block1, above of= W20, text width=2 cm,align=center, minimum height=2em] {Q-mBERT};
     \node (W31) [block1, above of= W30, text width=2 cm,align=center, minimum height=2em] {Q-mBERT};
     \node (W12) [block1, above of = W11, text width=2 cm,align=center, minimum height=2em] {$\bm s_{q}$};
     \node (W22) [block1, above of = W21, text width=2 cm,align=center, minimum height=2em] {$\bm s_{u_p}$};
     \node (W32) [block1, above of = W31, text width=2 cm,align=center, minimum height=2em] {$\bm s_{u_n}$};
     \node (W40) [block1, above of = W22, text width=4 cm,align=center, minimum height=2em] {Triplet Loss$(\bm s_q, \bm s_{u_p},\bm s_{u_n})$};

    \draw [thick,->,>=stealth] (W10) -- (W11);
    \draw [thick,->,>=stealth] (W11) -- (W12);
    \draw [thick,->,>=stealth] (W12) -- (W40);
    \draw [thick,->,>=stealth] (W20) -- (W21);
    \draw [thick,->,>=stealth] (W21) -- (W22);
    \draw [thick,->,>=stealth] (W22) -- (W40);
    \draw [thick,->,>=stealth] (W30) -- (W31);
    \draw [thick,->,>=stealth] (W31) -- (W32);
    \draw [thick,->,>=stealth] (W32) -- (W40);

    \end{tikzpicture}
    }
    \caption{Training procedure for relevance ranker teacher model.}
    \label{fig:relevance_ranker_training}
\end{figure}
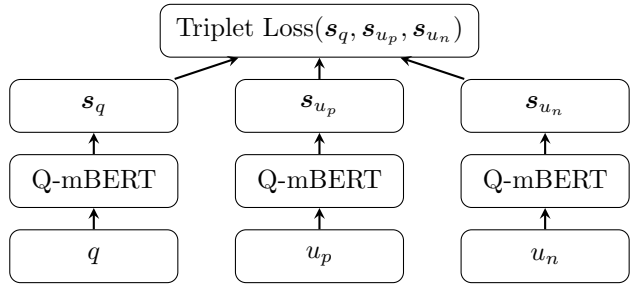

\tikzstyle{block} = [rectangle, draw, fill=blue!20, 
    text width=5em, text centered, rounded corners, minimum height=5em]
\tikzstyle{block1} = [rectangle, draw,text width=5em, text centered, rounded corners, minimum height=3em]

\newcommand*{\connectorH}[4][]{
  \draw[#1] (#3) -| ($(#3) !#2! (#4)$) |- (#4);
}
\newcommand*{\connectorV}[4][]{
  \draw[#1] (#3) |- ($(#3) !#2! (#4)$) -| (#4);
}

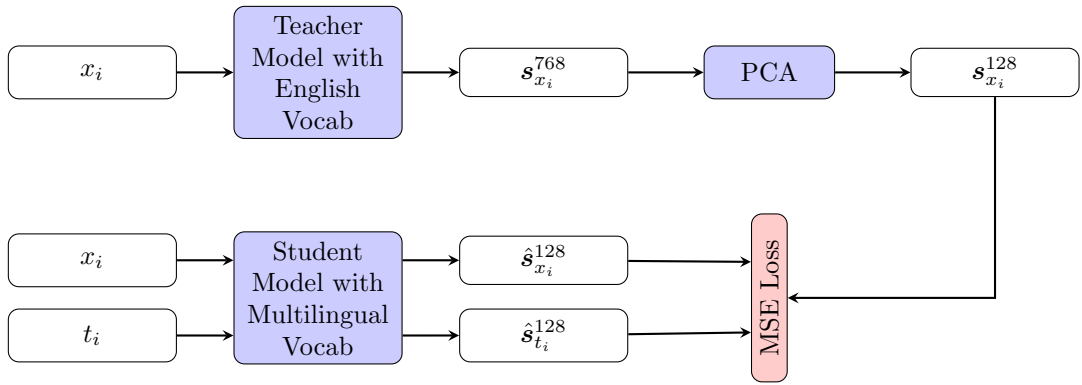
\begin{figure*}[t]
    \centering
    {
    \begin{tikzpicture}
     \node (W10) [block1, text width=2 cm,align=center, minimum height=2em] {$x_i$};
    \node (W11) [block,right of =W10, xshift = 2cm, text width=2 cm,align=center] {Teacher Model with English Vocab};
    \node (W21) [block, below of = W11, yshift = -2cm, text width=2 cm,align=center] {Student Model with Multilingual Vocab};
    \node (W12) [block1, right of = W11, xshift = 2 cm, text width=2 cm,align=center, minimum height=1em] {$\bm s^{768}_{x_i}$};
    \node (W13) [block, right of = W12, xshift = 2 cm, text width=1.5 cm,align=center, minimum height=2em] {PCA};
    \node (W20) [block1, text width=2 cm, align=center, minimum height=2em, xshift = -2cm, yshift = 0.5 cm, left of=W21] {$x_i$};
     \node (W30) [block1, text width=2 cm, align=center, minimum height=2em, below of=W20] {$t_i$};
     
     \node (W14) [block1, right of = W13, xshift = 2 cm, text width=2 cm,align=center,minimum height=1em] {$\bm s^{128}_{x_i}$};
     
     \node (W22) [block1, right of = W21, xshift = 2 cm, yshift = 0.5 cm, text width=2 cm,align=center,minimum height=1em] {$\hat{\bm s}^{128}_{x_i}$};
     
     \node (W23) [block1, below of = W22, text width=2 cm, align=center, minimum height=2em] {$\hat{\bm s}^{128}_{t_i}$};
     \node (W31) [block1, rotate=90, right of = W22, yshift = -3cm, xshift = -1.5 cm, ,fill=red!20,, text width=2 cm,align=center,minimum height=1em] {MSE Loss};
    
     \draw [thick,->,>=stealth] (W10) -- (W11);
     \draw [thick,->,>=stealth] (W11)--(W12);
     \draw [thick,->,>=stealth] (W12)--(W13);
     \draw [thick,->,>=stealth] (W13)--(W14);
     \draw [thick,->,>=stealth] (W21.24)--(W22);
     \draw [thick,->,>=stealth] (W21.336)--(W23.west);
     \draw [thick,->,>=stealth] (W23)--(W31.152);
     \draw [thick,->,>=stealth] (W22)--(W31.27);
     \draw [thick,->,>=stealth] (W20)--(W21.156);
     \draw [thick,->,>=stealth] (W30)--(W21.204);
    \draw [thick,->,>=stealth] (W14)|-(W31);
     
    \end{tikzpicture}
    }
    \caption{MSE Loss Calculation in a training step of S-mBERT Model with Embedding Vector Size of $1\times 128$ using a Parallel Sentence Pair $(x_i, t_i)$. The superscript for embedding vectors indicates its dimension, e.\,g. $\bm s^{768}_{x_i}$ has a dimension of $1\times 768$.}
    \label{fig:S-BERTDistillation}
\end{figure*}

\subsection{Teacher models}
As discussed in Section \ref{sec:search_system_design}, the aim of ranker modules in search engines is to find vector representations of documents in $\mathcal{U}$ that can be compared with the user query $q$ using a relevance ranker model and with the vector representations of previously stored user preferences using a semantic ranker model. There are large, open-sourced ranker models for these tasks which we use as teacher models for distillation.

We use the model paraphrase-xlm-r-multilingual-v1 \cite{reimer} as teacher semantic ranker model, trained on the Semantic Text Similarity (STS) \cite{Cer2018UniversalSE} dataset. The dataset lists pairs of sentences with labels describing the degree to which two sentences are semantically equivalent to each other. The model is trained using mean squared error loss for the cosine similarity between the sentence embedding and the gold label provided in the STS dataset, as shown in Fig.\,\ref{fig:semantic_ranker_training}

We used msmarco-distilroberta-base-v2 \cite{reimer1} as teacher relevance ranker model. It is trained using the MSMarco dataset \cite{Campos2016MSMA}, a set of user queries and passages. There are subsets of passages  that are relevant to each query. The relevance ranker teacher model is also trained using a siamese network with two key differences. Instead of sentence pairs, a triplet of query, relevant passage and irrelevant passage is used as input. The objective function is the triplet loss function where~--~given a query $q$, related documents $u_p$, and irrelevant documents $u_n$~--~it tunes the models to reduce the distance between the sentence embedding of $q$ and $u_p$ while also increasing the distance between $q$ and $u_n$. Mathematically, the loss function is to minimize

\begin{equation}
max(\lvert\lvert \bm s_q - \bm s_{u_p}  \lvert\lvert -   \lvert\lvert \bm s_q - \bm s_{u_n}   \lvert\lvert + \epsilon, 0)
\end{equation}
with $\bm s_x$ the sentence embedding of $q$, $u_p$, and $u_n$ (respectively) and $\lvert\lvert . \lvert\lvert $ the distance metric. The margin $\epsilon$ ensures that $\bm s_{u_p}$ is at least $\epsilon$ closer to $\bm s_q$ than $\bm s_{u_n}$.  

From the description of the teacher models we can deduce that the semantic ranker teacher model calculates the sentence embeddings in an embedding space such that semantically similar sentences are near to each other and vice versa. The relevance ranker teacher model maps the user query embedding closer to the relevant document embeddings than non-relevant document embeddings. The aim of the distillation procedure is to mimic these properties in the respective student models for the multilingual setting.

\section{Intuition of our proposed solution} \label{sec:proposed_solution}
As described in earlier sections, the distilled SBERT is suitable for efficient semantic and contextual ranking compared to the BERT model. Our proposed extensions of their distillation procedures reduce the required computational resources by several orders of magnitude.

We describe the architecture of the SBERT model briefly for understanding the intuition behind our proposed method. The SBERT model comprises multiple modules: a word-embedding layer, encoder layers, a pooling layer, and cosine similarity computation. 

The word-embedding layer is a map of tokens/words from the vocabulary and corresponding vectors of real numbers. These vectors are found by embedding algorithms such as word2vec \cite{Mikolov2013EfficientEO}. Thus, the word-embedding layer converts input text into a set of tokens from the vocabulary and converts that into a list of vectors. 

These lists of vectors are processed by multiple encoder layers comprising of self-attention and feed-forward layers. These processed vectors are then averaged by a pooling layer to find a representation of the input text in the form of a single vector. This final representation of different sentences can be compared to each other based on cosine similarity.  The training procedure for SBERT model for semantic and contextual rankers is described in Section \ref{sec:search_system_design}. 

An improvement of distillation is possible only in the word-embedding layer and encoder layers of the SBERT model since the other modules perform small computations. 
The word-embedding layer occupies most of the memory as large number of vectors, equal to the size of the vocabulary, must be stored. The encoder layers are responsible for most of the computations. With the reduction of the encoder layers, the performance of the SBERT model decreases drastically even after distillation.

Therefore, we propose that, by improving/increasing vocabulary especially in the multilingual setting, we can get performance for smaller models comparable to that of larger models. However, the increase in vocabulary increases the size of the word-embedding layer. We achieve small sized models by reducing the vector sizes in word-embedding layer with dimensionality reduction techniques such as PCA. The small vector size for word-embedding means that the final vector obtained from the SBERT model is also a small vector. This reduces the memory requirement for storing the user preferences as well.

\section{Extensions to SBERT Distillation}
\label{sec:distillation_extensions}
\label{SBERT}
This section describes the extension to the distillation procedure of SBERT model described in \cite{Reimers2020MakingMS}. 

\subsection{Generation of Multilingual Vocabulary } \label{sec:generate_vocab}
Many transformer-based language models have been developed and open-sourced. Most of them are monolingual and only some are bilingual \cite{Lample2019CrosslingualLM} or  multilingual \cite{Conneau2020UnsupervisedCR}, \cite{Liu2020MultilingualDP}, \cite{Pires2019HowMI}.  Multilingual models are trained on data for hundreds of languages. The vocabularies for these open-source models are monolingual, bilingual or multilingual. 

There is no readily available solution for generating a vocabulary for a set of target languages. Therefore, the multilingual SBERT model uses the available vocabulary of hundreds of languages for the distillation. This is inefficient if a student model must be trained only for few languages, as many of the tokens used in the multilingual vocabulary would be unused~--~making it equivalent to a very small vocabulary size. 

We propose to generate a wordpiece \cite{Wu2016GooglesNM} vocabulary with the data only for the target languages. We used Wikipedia dumps for generating this vocabulary. As the size of Wikipedia dumps varies greatly for different languages, the low-resource languages are under-represented in the vocabulary. 

To mitigate this problem, we performed exponential smoothed weighting of data. The percentage of data for each language is exponentiated with a factor (we used 0.7) and renormalised to get a probability distribution used for sampling the data for vocabulary creation. This helps in under-sampling of high-resource languages and over-sampling of low-resource languages.

\begin{table*}[t]
\centering
 \begin{tabular}{|p{0.1\textwidth}|  p{0.05\textwidth}| p{0.1\textwidth}| p{0.05\textwidth}| p{0.05\textwidth}| p{0.05\textwidth}| p{0.05\textwidth}| p{0.05\textwidth}| p{0.05\textwidth} |p{0.05\textwidth} | p{0.05\textwidth} |p{0.05\textwidth}|} 
 \hline
 Model Name & Size (MB) & Vocabulary Size &  ES-ES & ES-EN & EN-EN & TR-EN & EN-DE & FR-EN & IT-EN & NL-EN & Avg.  \\ [0.5ex] 
 \hline
 
  paraphrase-xlm-r-multilingual-v1 ($12/768$) & 1100 & 250002 &  85.7 & 84.1 & 88.1	& 80.2	& 83.7	& 83.2	& 85.7	& 83.7 & 84.3  \\
   \hline
   
   paraphrase-xlm-r-multilingual-v1 ($12/128$) & 1100  & 250002  & 85.0 & 83.7 & 87.2	& 79.34	& 83.0	& 82.2	& 84.8	& 83.0 & 83.53  \\
   \hline
   S-mBERT ($1/768$)  & 312 & 100000 & 80.64 &	66.52 &	84.05 &	59.56 &	67.20 &	68.58 &	65.79 &	69.04 &  70.17 \\
 \hline
  S-mBERT ($1/128$)  & 27 & 50000 & 78.61 &	60.07 &	82.60 &	52.54 &	63.93 &	63.58 &	60.48 &	63.82 &  65.7 \\
 \hline
  
   S-mBERT ($1/128$)  & 53 & 100000 & 79.91 &	68.49 &	83.90 &	58.91 &	63.93 &	68.59 &	67.74 &	69.91 & 70.17  \\
 \hline
   
   S-mBERT ($1/128$)  & 79 & 150000 & 80.56 &	62.37 &	83.24 &	60.16 &	62.02 &	61.65 &	59.24 &	64.01 & 66.65  \\
 \hline

   S-mBERT ($1/128$)  & 56 & 105879 (multilingual BERT vocabulary) & 81.40 &	63.11 &	82.24 &	57.68 &	62.76 &	62.82 &	59.72 &	65.45 & 66.89 \\
 \hline
 \end{tabular}
 \caption{Spearman rank correlation $\rho$ for teacher and student semantic ranker models using the STS-Extended Dataset \cite{Reimers2020MakingMS}. The model architecture is described in the form of (\# encoder layers/embedding dimension) written afteer the model names.}
 \label{Table_XaynBERTEvaluation}
\end{table*}

\begin{table*}[h!]
\centering
 \begin{tabular}{|p{0.2\textwidth} |p{0.1\textwidth}| p{0.1\textwidth}| p{0.1\textwidth}| p{0.1\textwidth}| p{0.1\textwidth}|} 
 \hline
 Model Name & Size (MB) & Vocabulary Size & MRR@10 & NDCG@10 & MAP@100  \\
  \hline 
  msmarco-distilroberta-base-v2 ($6/768$) & 315 & 50265 & 0.5874 & 0.6284 & 0.5868 \\
 \hline 
  msmarco-distilroberta-base-v2 (($6/768$)) & 315 & 50265 & 0.5678  & 0.6088 & 0.5676 \\
 \hline
  QA-mBERT ($1/128$) & 27 & 50000 & 0.2539 & 0.2841 & 0.2576   \\
 \hline
  QA-mBERT ($1/128$) & 53 & 100000 & 0.2904 & 0.3226 & 0.2931   \\
 \hline
  QA-mBERT ($1/128$) & 79 & 150000 & 0.2832 & 0.3140 & 0.2861   \\
 \hline
  QA-mBERT ($1/128$) & 312 & 100000 & 0.3119 & 0.3455 & 0.3151   \\
 \hline
  QA-mBERT ($1/128$) & 56 & 105879 (multilingual BERT vocabulary) & 0.2408 & 0.2688 & 0.2438  \\ 
 \hline
 \end{tabular}
 \caption{Model comparison on 6980 queries on the MSMARCO test dataset with a maximum corpus size of 1 million and cosine similarity scoring function. The model architecture is described in the form of (\# encoder layers/embedding dimension) besides the model names.}
 \label{Table_qbert_results}
\end{table*}

\begin{table*}[h!]
\centering
 \begin{tabular}{|p{0.25\textwidth}|p{0.25\textwidth}|p{0.25\textwidth}|} 
 \hline
 Model Name  & Energy Consumption [kj] & Inference Time [ms] \\
  \hline
  paraphrase-xlm-r-multilingual-v1 ($12/768$) & 2382.99 [2079.55, 2573.30] & 153.11 [143.16, 170.04]
 \\
   \hline
  paraphrase-xlm-r-multilingual-v1 ($12/128$) & 2760.83 [2582.50, 3144.49] & 172.30 [144.10, 182.46]
 \\
  \hline
  S-mBERT ($1/768$) & 350.77 [311.52, 377.53] & 6.81 [5.64, 7.34] \\
  \hline
  S-mBERT ($1/128$) & 120.12 [113.81, 173.29] & 2.34 [2.06, 2.58] \\
  \hline
  msmarco-distilroberta-base-v2 ($6/768$) & 1072.17 [949.06, 1160.15] & 72.05 [60.29, 84.46] \\
  \hline
  msmarco-distilroberta-base-v2 ($6/128$) & 937.95 [899.68, 1039.65] & 75.78 [71.96, 91.39] \\
  \hline
    QA-mBERT ($1/768$) & 167.33 [147.39, 352.40]  & 9.56 [7.35, 12,79] \\
 \hline
  QA-mBERT ($1/128$) & 121.22 [111.92, 167.11] & 2.35 [2.08, 2.62] \\
  \hline
 \end{tabular}
 \caption{Median energy consumption (in kilojoule) and inference time (in milliseconds) reported with the first and third quartile. The model architecture is described in the form of (\# encoder layers/embedding dimension) besides the model names.}
 \label{Table_energyconsumption}
\end{table*}

\subsection{Dimensionality Reduction of Embedding Space}

To imitate the properties of sentence-transformer teacher model, we train a student model such that the embedding space is similar, in that the points are embedded in proximity to those embeddings of the teacher model. This ensures that the behavior of the cosine-similarity module is very similar when using our distilled student model.

To make the model multilingual, we map the translated sentence pairs. For this,  
we use the distillation procedure to
compress the S-BERT teacher models into a multilingual small student model. 
This requires a corpus of parallel sentences $((x_1, t_1),\ldots,(x_n, t_n))$, where $x_i$ is the sentence in the source language and $t_i$ is the sentence in the target language. 
The student model is trained such that $\hat{\bm s}_{x_i} \approx {\bm s}_{x_i}$ and $\hat{\bm s}_{t_i} \approx {\bm s}_{x_i}$. Specifically, for a given mini-batch $\beta$, the mean-squared loss (MSE) is calculated as 
\begin{equation}
MSE = \frac{1}{\lvert\beta\lvert}\sum_{i\in\beta}\left(\lvert\lvert\hat{\bm s}_{x_i} - {\bm s}_{x_i}\lvert\lvert^2_2 + \lvert\lvert\hat{\bm s}_{x_i} - {\bm s}_{x_i}\lvert\lvert^2_2\right).
\end{equation}

This loss function allows for a completely different architecture and vocabulary for the student model than those used for the teacher model. The only constraint for this is that $\hat{\bm s}_{x}$ and ${\bm s}_{x}$ have the same dimension. This enables us to generate our own vocabulary, as described in Section \ref{sec:generate_vocab}. 

We desire a smaller embedding dimension as this will reduce the size of the vocabulary and the size of the vector representations for user preferences. We use principal component analysis (PCA) for learning parameters of an additional layer on top of the teacher model to reduce the dimension of ${\bm s}_{x_i}$ to desired size. The entire distillation method is described in Fig.\,\ref{fig:S-BERTDistillation}. The distillation for both the semantic and the relevance ranker is same, the only difference being in the choice of the teacher model.

\section{Experimental results} \label{sec:comparative_analysis}

We distilled both semantic and relevance ranker student models using the Tatoeba dataset \cite{Artetxe2019MassivelyMS} for languages French, German, Italian, Spanish, Polish, Portugese, Dutch, Russian, Turkish. We distilled models for 20 epochs using MSE loss, a batch-size of 128, the ADAM optimizer with learning rate $2* 10^{-5}$, and a linear learning rate warm-up over 10\% of the training data. For semantic ranker, we used 100,000 sentence-pairs (English to target language) for each target languages while the for relevance ranker, we used 500,000 sentence pairs for each target language.

\subsection{Performance Measurement}
The evaluation of the semantic ranker models on the STS test dataset is given in Table \ref{Table_XaynBERTEvaluation}, showing the Spearman rank correlation $\rho$ between the cosine similarities of sentence representations and the gold labels for the STS extended dataset \cite{Reimers2020MakingMS}. We present the results for only those language pairs for which the STS extended dataset contains test examples. Also, our primary aim is to showcase the improvement that can be achieved by the proposed methods and not to achieve the state-of-the-art results. We can achieve better results with additional parallel sentence  datasets. By convention, the performance is reported with $\rho \times 100$. Our observations from these tables are:
\begin{enumerate}
\item The student model shows acceptable performance even after extreme compression of the teacher model from 1100 MB to ~27 MB with our extension of distillation.

\item Chosing an optimal vocabulary size is important: for the same model architecture, varying the vocabulary size provides different results. Increasing the size of generated vocabulary from 50,000 to 100,000 shows a ~4.5 points improvement in the STS score on average. Further increasing the vocabulary size decreases that score by roughly ~3.5 points on average. Using the vocabulary of the open-sourced multilingual BERT model results in a score about $3.28$ points lower than that of the distilled student model with the same architecture and similar size of vocabulary.

\item Distilling a student model with original dimension of the teacher model ($1 \times 768$) and distilling a student model with reduced dimension ($1 \times 128$) provides similar results on average but with significant decrease in the model size. 

\end{enumerate}

The evaluation of the relevance ranker model using the MSMarco test dataset is shown in Fig.\,\ref{Table_qbert_results}; we can see similar trends as observed with the semantic ranker student models, confirming the importance of our contributions of generating an optimal vocabulary and reductions of the embedding dimension of the teacher model.

\subsection{Computational Resources Measurement}
Our experiments demonstrate that the distillation method we presented in this paper reduces the size of the ranker models significantly. To provide evidence for the efficiency of our distilled models, we compared them to the respective teacher models against two other important criteria for edge devices, energy consumption and inference speed (see Table~\ref{Table_XaynBERTEvaluation}). 

All our experiments were conducted on a Lenovo ThinkPad X1 Carbon with Intel Core i7 CPU processor with max clock speed of 2.09 GHz and 16 GB RAM using python library pyJoules \cite{pyJoulesLibrary}. We tracked energy consumption of models for 1000 runs. The results of the energy consumption of all teacher and student models are summarised in Table \ref{Table_energyconsumption}. These results show that the best student models obtained through the distillation procedure developed in this paper require significantly less energy and inference time.

\section{Conclusion}
\label{sec:conclusion}
In this paper, we proposed and developed two extensions to the distillation procedure of sentence-transformer ranker models: generation of an optimal size vocabulary and dimensionality reduction
of the embedding dimension of teachers prior to distillation. 

We applied these extensions to a known distillation procedure. We showed experimentally that applying this extended distillation procedure to semantic and relevance ranker models resulted in extreme compression that retains desired performance, increased inference speed, and reduced energy consumption significantly.

In future work, we plan to apply and evaluate additional model compression techniques such as quantization to the distilled ranker models. 

We will use the distilled ranker models developed in this paper and their enrichment with further compression techniques in a personalised, private search engine that we currently design and implement for use on edge devices. 

{\small
\bibliographystyle{ieee_fullname}
\bibliography{PaperForReview}

\begin{thebibliography}{10}\itemsep=-1pt

\bibitem{reimer1}
sentence-transformers/msmarco-distilroberta-base-v2 · hugging face.

\bibitem{Artetxe2019MassivelyMS}
Mikel Artetxe and Holger Schwenk.
\newblock Massively multilingual sentence embeddings for zero-shot
  cross-lingual transfer and beyond.
\newblock {\em Transactions of the Association for Computational Linguistics},
  7:597--610, 2019.

\bibitem{Brown2020LanguageMA}
Tom~B. Brown, Benjamin Mann, Nick Ryder, Melanie Subbiah, Jared Kaplan,
  Prafulla Dhariwal, Arvind Neelakantan, Pranav Shyam, Girish Sastry, Amanda
  Askell, Sandhini Agarwal, Ariel Herbert-Voss, Gretchen Krueger, T.~J.
  Henighan, Rewon Child, Aditya Ramesh, Daniel~M. Ziegler, Jeff Wu, Clemens
  Winter, Christopher Hesse, Mark Chen, Eric Sigler, Mateusz Litwin, Scott
  Gray, Benjamin Chess, Jack Clark, Christopher Berner, Sam McCandlish, Alec
  Radford, Ilya Sutskever, and Dario Amodei.
\newblock Language models are few-shot learners.
\newblock {\em ArXiv}, abs/2005.14165, 2020.

\bibitem{Campos2016MSMA}
Daniel~Fernando Campos, Tri Nguyen, Mir Rosenberg, Xia Song, Jianfeng Gao,
  Saurabh Tiwary, Rangan Majumder, Li Deng, and Bhaskar Mitra.
\newblock Ms marco: A human generated machine reading comprehension dataset.
\newblock {\em ArXiv}, abs/1611.09268, 2016.

\bibitem{Cer2018UniversalSE}
Daniel~Matthew Cer, Yinfei Yang, Sheng yi Kong, Nan Hua, Nicole Limtiaco,
  Rhomni~St. John, Noah Constant, Mario Guajardo-Cespedes, Steve Yuan, Chris
  Tar, Yun-Hsuan Sung, Brian Strope, and Ray Kurzweil.
\newblock Universal sentence encoder.
\newblock {\em ArXiv}, abs/1803.11175, 2018.

\bibitem{Conneau2020UnsupervisedCR}
Alexis Conneau, Kartikay Khandelwal, Naman Goyal, Vishrav Chaudhary, Guillaume
  Wenzek, Francisco Guzm{\'a}n, Edouard Grave, Myle Ott, Luke Zettlemoyer, and
  Veselin Stoyanov.
\newblock Unsupervised cross-lingual representation learning at scale.
\newblock In {\em ACL}, 2020.

\bibitem{pyJoulesLibrary}
Arthur d'Azémar, Romain Rouvoy, Chakib Belgaid, Alex Kaminetzky, and Benjamin
  Danglot.
\newblock pyjoules.
\newblock \url{https://github.com/powerapi-ng/pyJoules}, 2019.

\bibitem{devlin2018bert}
Jacob Devlin, Ming-Wei Chang, Kenton Lee, and Kristina Toutanova.
\newblock Bert: Pre-training of deep bidirectional transformers for language
  understanding.
\newblock {\em arXiv preprint arXiv:1810.04805}, 2018.

\bibitem{Gholami2022ASO}
Amir Gholami, Sehoon Kim, Zhen Dong, Zhewei Yao, Michael~W. Mahoney, and Kurt
  Keutzer.
\newblock A survey of quantization methods for efficient neural network
  inference.
\newblock {\em ArXiv}, abs/2103.13630, 2022.

\bibitem{Hinton2015DistillingTK}
Geoffrey~E. Hinton, Oriol Vinyals, and Jeffrey Dean.
\newblock Distilling the knowledge in a neural network.
\newblock {\em ArXiv}, abs/1503.02531, 2015.

\bibitem{Jiao2020TinyBERTDB}
Xiaoqi Jiao, Yichun Yin, Lifeng Shang, Xin Jiang, Xiao Chen, Linlin Li, Fang
  Wang, and Qun Liu.
\newblock Tinybert: Distilling bert for natural language understanding.
\newblock {\em ArXiv}, abs/1909.10351, 2020.

\bibitem{Lample2019CrosslingualLM}
Guillaume Lample and Alexis Conneau.
\newblock Cross-lingual language model pretraining.
\newblock In {\em NeurIPS}, 2019.

\bibitem{Liu2020MultilingualDP}
Yinhan Liu, Jiatao Gu, Naman Goyal, Xian Li, Sergey Edunov, Marjan
  Ghazvininejad, Mike Lewis, and Luke Zettlemoyer.
\newblock Multilingual denoising pre-training for neural machine translation.
\newblock {\em Transactions of the Association for Computational Linguistics},
  8:726--742, 2020.

\bibitem{Liu2019RoBERTaAR}
Yinhan Liu, Myle Ott, Naman Goyal, Jingfei Du, Mandar Joshi, Danqi Chen, Omer
  Levy, Mike Lewis, Luke Zettlemoyer, and Veselin Stoyanov.
\newblock Roberta: A robustly optimized bert pretraining approach.
\newblock {\em ArXiv}, abs/1907.11692, 2019.

\bibitem{Mikolov2013EfficientEO}
Tomas Mikolov, Kai Chen, Gregory~S. Corrado, and Jeffrey Dean.
\newblock Efficient estimation of word representations in vector space.
\newblock In {\em ICLR}, 2013.

\bibitem{Pires2019HowMI}
Telmo J.~P. Pires, Eva Schlinger, and Dan Garrette.
\newblock How multilingual is multilingual bert?
\newblock In {\em ACL}, 2019.

\bibitem{248452}
R. Reed.
\newblock Pruning algorithms-a survey.
\newblock {\em IEEE Transactions on Neural Networks}, 4(5):740--747, 1993.

\bibitem{reimer}
Nils Reimer.
\newblock Sentence-transformers/paraphrase-xlm-r-multilingual-v1 · hugging
  face.

\bibitem{Reimers2019SentenceBERTSE}
Nils Reimers and Iryna Gurevych.
\newblock Sentence-bert: Sentence embeddings using siamese bert-networks.
\newblock {\em ArXiv}, abs/1908.10084, 2019.

\bibitem{Reimers2020MakingMS}
Nils Reimers and Iryna Gurevych.
\newblock Making monolingual sentence embeddings multilingual using knowledge
  distillation.
\newblock In {\em EMNLP}, 2020.

\bibitem{Sanh2019DistilBERTAD}
Victor Sanh, Lysandre Debut, Julien Chaumond, and Thomas Wolf.
\newblock Distilbert, a distilled version of bert: smaller, faster, cheaper and
  lighter.
\newblock {\em ArXiv}, abs/1910.01108, 2019.

\bibitem{simon_2021}
Julien Simon.
\newblock Large language models: A new moore's law?, Oct 2021.

\bibitem{Sun2020MobileBERTAC}
Zhiqing Sun, Hongkun Yu, Xiaodan Song, Renjie Liu, Yiming Yang, and Denny Zhou.
\newblock Mobilebert: a compact task-agnostic bert for resource-limited
  devices.
\newblock {\em ArXiv}, abs/2004.02984, 2020.

\bibitem{Wu2016GooglesNM}
Yonghui Wu, Mike Schuster, Z. Chen, Quoc~V. Le, Mohammad Norouzi, Wolfgang
  Macherey, Maxim Krikun, Yuan Cao, Qin Gao, Klaus Macherey, Jeff Klingner,
  Apurva Shah, Melvin Johnson, Xiaobing Liu, Lukasz Kaiser, Stephan Gouws,
  Yoshikiyo Kato, Taku Kudo, Hideto Kazawa, Keith Stevens, George Kurian,
  Nishant Patil, Wei Wang, Cliff Young, Jason~R. Smith, Jason Riesa, Alex
  Rudnick, Oriol Vinyals, Gregory~S. Corrado, Macduff Hughes, and Jeffrey Dean.
\newblock Google's neural machine translation system: Bridging the gap between
  human and machine translation.
\newblock {\em ArXiv}, abs/1609.08144, 2016.

\bibitem{Zhao2019ExtremeLM}
Sanqiang Zhao, Raghav Gupta, Yang Song, and Denny Zhou.
\newblock Extreme language model compression with optimal subwords and shared
  projections.
\newblock {\em ArXiv}, abs/1909.11687, 2019.

\end{thebibliography}

}

\end{document}